# PATTERN RECOGNITION USING INVERSE RESONANCE FILTRATION

Olga Sofina[1], Yuriy Bunyak[2], Roman Kvetnyy[3]

[1] Vinnitsa National Technical University. 95 Khmelnitske sh. Vinnitsa 21021 Ukraine. olya_bunyak@mail.ru
[2] InnoVinn Inc. 14 Kievskaya str. Vinnitsa 21001 Ukraineinn. yuri.bunyak@innovinn.com
[3] Vinnitsa National Technical University. 95 Khmelnitske sh. Vinnitsa 21021 Ukraine. rkvetny@mail.ru

**ABSTRACT**
An approach to textures pattern recognition based on inverse resonance filtration (IRF) is considered. A set of principal resonance harmonics of textured image signal fluctuations eigen harmonic decomposition (EHD) is used for the IRF design. It was shown that EHD is invariant to textured image linear shift. The recognition of texture is made by transfer of its signal into unstructured signal which simple statistical parameters can be used for texture pattern recognition. Anomalous variations of this signal point on foreign objects. Two methods of 2D EHD parameters estimation are considered with the account of texture signal breaks presence. The first method is based on the linear symmetry model that is not sensitive to signal phase jumps. The condition of characteristic polynomial symmetry provides the model stationarity and periodicity. Second method is based on the eigenvalues problem of matrices pencil projection into principal vectors space of singular values decomposition (SVD) of 2D correlation matrix. Two methods of classification of retrieval from textured image foreign objects are offered.

## 1. Introduction

Textured images pattern recognition is one of the main problems in visual basing measurement and control systems. Main methods of texture analysis and modeling are presented in reviews [1-3]. Most of the methods are based on statistical analysis, spectral transforms and dynamic models of images signal. Method choosing depends on the texture type – regular, quasi-regular, stochastic or dynamic.

The main purpose of the texture analysis consists in obtaining the minimal number parameters that are invariant to spatial and temporal image transforms. The models of linear and nonlinear autoregression, autoregression and moving average can represent correlation relation of texture structure. The correlation can be used directly too, the most known is the method of co-occurrence matrices. Another approach is based on the images signal spectral transforms (ST) into spatial frequency domain where spectral parameters of the image have simple statistic in narrow range [4,5]. The newest methods combine statistical models and ST using a set of some type functions. The most popular became the wavelet functions and their generalizations in the case of 2D transforms – ridgelets and curvelets [6-12]. The wavelets represent well a local geometry feature of textures with minimal number of significant coefficients however they do not represent as well the textures periodicity because are intended for presentation of transient signals. Aspiration for textures presentation by minimal number of parameters causes application of new ST especially in combination with known. The Radon transform joined with usual and wavelet based ST gives the possibility to present images in frequency-angle domain that has a much lower dimension because it is scale- and affine-invariant [12-14]. The eigenvector decomposition (EVD) of texture kernel and filters that are resonance to some type of image eigenmodes are most informative when texture is dynamic [15,16]. The EVD is computationally complicate and therefore independent component analysis (ICA), empirical mode decomposition (EMD) and SVD are using instead of it [17-21]. However, the decompositions usually are using for projection transform of full image space that complicates their application in real time systems. Another way is based on using of filters banks. The filters banks are synthesized with the help of Fourier transform, wavelets and Gabor functions [1-3,21]. The wavelets and Gabor functions have parameters that are polysemantic and unique technique for their estimation according to texture structure does not exist. This complicates the problem of filters banks design. In the case of dynamic textures the decompositions are used for synthesis of texture model which generates an original like texture [22,23]. The textures correlation matrix often serves as object of the decompositions because it represents dynamic properties of large image in compact form. The asymptotic correlation matrix of a stationary process has Toeplitz structure and its eigenvectors have harmonic nature [24]. Therefore the harmonic decomposition is natural approximation of EVD or EMD. The parameters of harmonics can be agreed with periods of texture spatial structure. The planar and spherical eigen harmonic decompositions (EHD) were used for compact presentation of textured and face like images [25-27].

The brief survey of the textured images analysis methods allows to make the following conclusions. The most informative is decomposition that is invariant to texture spatial and temporal dynamic. The wavelets and SVD ignore signal structural information because are not shift invariant. The EVD and EMD have not analytical definition that complicates their use for approximation and interpolation of the image fragments. In contrast to mentioned decompositions, the EHD has simple analytical definition and joins capabilities for approximation and interpolation of 2D fields [28]. And so the EHD is appropriate functions basis for approximation of textured image fragment and for interpolation of its pattern for comparative analysis of rest image part. Another way of this procedure implementation is based on suppression of the texture pattern by EHD and obtaining of approximately flat signal. The signal flatness level points on recognized pattern or on the foreign object. Such spectral transform can be realized in spatial domain by convolution scheme which can be defined as inverse resonance filter (IRF).

## 2. Harmonic Model of Textured Image

Ideal textured image of size $N \cdot P \times M \cdot Q$ can be represented as tensor product of two matrices,

$$\mathbf{X} = \mathbf{T} \otimes \mathbf{B}, \quad (1)$$

where matrix $\mathbf{T}$ of size $N \times M$ includes unit elements with insignificant fluctuations and matrix $\mathbf{B}$ of size $P \times Q$ is the texture kernel, $\otimes$ – tensor product. Let's the texture (1) is changing by moving of the matrix $\mathbf{B}$ along $t$ rows and $\tau$ columns. The model of such shift is the following: rows (columns) of the matrix move cyclically but instead last row (column), which becomes first, new row (column) is introduced as linear combination of previous ones. This shift is presented by the operator

$$\mathbf{K}_P = \begin{bmatrix} 0 & 1 & \cdots & 0 \\ \vdots & \vdots & \ddots & \vdots \\ 0 & 0 & \cdots & 1 \\ -a_P & -a_{P-1} & \cdots & -a_1 \end{bmatrix} \quad (2)$$

that can be called as linear shift operator (LSO), its characteristic polynomial

$$1 + \sum_{i=1}^{P} a_i z^i = 0 \quad (3)$$

has the roots $z_i = \exp(i 2\pi f_i)$, $f_i$ – eigen or resonance frequencies, $i = 1 \ldots P$. The LSO (2) can be defined for two image coordinates. The spectral factorization of the LSO can be written for both coordinates as

$$\mathbf{K}_{xP}^t = \mathbf{Z}_{xP} \mathbf{diag}\left[z_{xi}^t\right]_{i=1 \ldots P} \mathbf{Z}_{xP}^{\#};$$
$$\mathbf{K}_{yQ}^t = \mathbf{Z}_{yQ} \mathbf{diag}\left[z_{yi}^t\right]_{i=1 \ldots Q} \mathbf{Z}_{yQ}^{\#}, \quad (4)$$

where $\mathbf{Z}_{xP} = \left[z_{xi}^t\right]_{j=1 \ldots P}^{t=0 \ldots P-1}$, $\mathbf{Z}_{yQ} = \left[z_{yi}^t\right]_{i=1 \ldots Q}^{t=0 \ldots Q-1}$, $\#$ – pseudoinversion. The linear shifted kernel matrix can be presented using (4) in the manner

$$\mathbf{B}^{t,\tau} = \mathbf{K}_{xP}^t \mathbf{B} \mathbf{K}_{yQ}^{\tau T} =$$

$$= \mathbf{Z}_{xP} \mathbf{diag}\left[z_{xi}^t\right]_{j=1 \ldots P} \mathbf{Z}_{xP}^{\#} \mathbf{B} \mathbf{Z}_{yQ}^{\#T} \mathbf{diag}\left[z_{yi}^\tau\right]_{i=1 \ldots Q} \mathbf{Z}_{yQ}^T \quad (5)$$

$$= \mathbf{Z}_{xP} \mathbf{diag}\left[z_{xi}^t\right]_{i=1 \ldots P} \mathbf{A} \ \mathbf{diag}\left[z_{yi}^\tau\right]_{i=1 \ldots Q} \mathbf{Z}_{yQ}^T,$$

where

$$\mathbf{A} = \mathbf{Z}_{xP}^{\#} \mathbf{B} \mathbf{Z}_{yQ}^{\#T} \quad (6)$$

– spectrum of the matrix $\mathbf{B}$ in the basis $\mathbf{Z}$, $T$ – transposition. Expression (5) shows the spectrum invariance to LSO – spectral matrix of the shifted kernel in (5) differs from the original one by phase multiplies. The periodicity of kernel in textured image is defined by condition

$$\mathbf{B}^{mP, nQ} = \mathbf{B}, \quad (7)$$

where $m$ and $n$ arbitrary integer values. This condition allows to present the full ideal textured image (1) by overdetermined matrices.

$$\mathbf{X} \approx \widetilde{\mathbf{Z}}_{xP} \mathbf{A} \ \widetilde{\mathbf{Z}}_{yQ}^T, \quad (8)$$

where

$$\widetilde{\mathbf{Z}}_{xP} = \left[z_{xi}^t\right]_{j=1 \ldots P}^{t=0 \ldots NP-1}, \ \widetilde{\mathbf{Z}}_{yQ} = \left[z_{yi}^t\right]_{i=1 \ldots Q}^{t=0 \ldots MQ-1}.$$

The expression (8) has approximate invariance to linear shift transforms because the pseudoinversion of overdetermined matrices is approximative. The linear shift is natural for many models of texture, static and dynamic. And so the EHD (8) provides informative presentation of the image model on condition the parameters of the LSO are estimated correspondingly to texture structure.

## 3. Methods of Harmonic Decomposition and Textures Filtering

The problem of texture filtering and recognizing can be considered as coherent suppression of its structure. The filtered signal must be flat as possible with the additive error noise. The flatness level is characterized by noise dispersion value. The filtration is based on two transforms (6), (8) using square or overdetermined matrices $\mathbf{Z}$ and modification of the spectral matrix $\mathbf{A}$. The first step of the filter design is estimation of the LSO (2) parameters and definition of its characteristic polynomial (3) roots. There is known a series of methods for estimation of 2D data resonance frequencies [24]. In the case of texture EHD it is necessary to take into account important restrictions to model parameters:
- The texture model is considered as stationary process with approximative pattern periodicity (7). As it

follows from condition (7) the roots of characteristic polynomial (3) must be placed on the unit circle.
- Texture can be not smooth 2D function and the method must be not-sensitive to signal breaks.
- The order of the LSO must corresponds to texture structure – the matrix **A** must not includes zero components. But if the EHD order is understated the spectrum will be corrupted.
- The high order filter is required when texture is quasi-regular, dynamic or corrupted by noise spikes.

**3.1. The Linear Symmetry Model Based Method**

As it was shown in [28] the eigen harmonics of two one-dimensional data vectors along coordinates can be used for 2D data representation by EHD approximation (11). Therefore usual methods of harmonic decomposition can be used for LSO parameters estimation [24]. The methods of spectral parameters estimation are based on the idea of correlation relation between two mutually shifted data sets. The linear parameterization of this relation yields the linear prediction (LP) model and its various implementations. The LP based methods are sensitive to signal phase, usually it becomes apparent in splitting of spectral lines. The image signal breaks can be considering as phase interruptions. We enforce the LP model by condition of invariance to shift operations with the purpose of reducing sensitivity to data breaks. This invariance may be determined as linear symmetry (LS) [29]. The LS of correlation matrix $\mathbf{R}_p$ of size $p \times p$ has the manner

$$\mathbf{R}_p = \mathbf{K}_p \mathbf{R}_p \mathbf{K}_p^H \qquad (9)$$

and it differs from usual LP as superposition of direct and conjugate ($H$ – Hermitian conjunction) LP that removes depending on signal phase. The equation (9) provides unitarity of LSO, the roots of polynomial (3) are placed on unit circle, but it is valid on condition of Toeplitz structure of correlation matrix. Therefore the exact LS (9) can be changed by the approximative condition of LS that is optimized accordingly to maximal likelihood criteria,

$$\frac{\partial}{\partial a_i} tr(\mathbf{R}_p^{-1} \mathbf{K}_p \mathbf{R}_p \mathbf{K}_p^H) = 0; \quad i = 1,\ldots, p \,. (10)$$

The equation (10) has simple solution

$$a_{p-i} = -\rho_{i,p} \rho_{p-1,p-1}^{-1}, \quad i = 1\ldots p-1; \quad a_p = 1,$$

where $\rho_{\cdot,p-1}$ are elements of the last column of the inverse matrix $\mathbf{R}_P^{-1}$, the dispersion of LS model error is estimated as $\sigma^2 = \rho_{p-1,p-1}^{-1}$. As it follows from this solution the LS model order $p$ must be chosen such that element $\rho_{p-1,p-1}$ is the maximum or local maximum of the function $\rho_{\cdot,p-1}$. In the case of LS (10) the condition of unitary symmetry can be achieved by additional relations between LSO parameters: $a_p = 1;\ a_i = a_{p-i},\ i = 1\ldots p/2 - 1$. Taking into account this polynomial symmetry the expression (25) can be written as equation system with respect to the polynomial coefficients.

$$\sum_{i=1}^{q-1} a_i (r_{i,k} + r_{p-i,k} + r_{i,p-k} + r_{p-i,p-k})$$
$$+ a_q (r_{k,q} + r_{p-k,q}) = \qquad (11a)$$

$$\rho_{p-1,p-1}^{-1} \sum_{i=0}^{p-2} \rho_{i,p-1} (r_{k,i+1} + r_{p-k,i+1}) - r_{k,0} - r_{p-k,0},$$

where $p$ is even and $k = 1\ldots q-1$, $q = p/2 - 1$, for $k = q$

$$\sum_{i=1}^{q} a_i (r_{i,q} + r_{P-i,q}) + a_q r_{q,q} =$$

$$\rho_{p-1,p-1}^{-1} \sum_{i=0}^{p-2} \rho_{p-1,i} r_{i+1,q} - r_{q,0}. \qquad (11b)$$

Analogues equation can also be written for the odd model order. But the case of even order is preferable because polynomial (3) gives $p/2$ pairs of mutually complex conjugated significant roots. In the case of odd order the equation (11) yields the polynomial (3) with one insignificant root, usually equal to constant value. The spectrum component of this root is undesirable for texture filtering because it is presented in filtered signal.

If image signal is homogeneous than the first data moments of some row and column can be used for estimation of LSO parameters. In other case the 2D correlation matrix

$$\mathbf{R}^{(2)} = \left[ r_{i_x,k_x,i_y,k_y}^{(2)} \right]_{i_x,k_x=0\ldots P-1}^{i_y,k_y=0\ldots Q-1} =$$
$$\left[ \sum_{m=0}^{n_x-P} \sum_{n=0}^{n_y-Q} u_{m+i_x,n+i_y} u_{m+k_x,n+k_y} \right]_{i_x,k_x=0\ldots P-1}^{i_y,k_y=0\ldots Q-1} \qquad (12)$$

is more informative The correlation matrices that characterize data along coordinates in (10) may be defined as

$$\mathbf{R}_x^{(2)} = \left[ r_{i_x,k_x,0,0}^{(2)} \right]_{i_x,k_x=0\ldots P-1}^{i_y,k_y=0} ;$$

$$\mathbf{R}_y^{(2)} = \left[ r_{0,0,i_y,k_y}^{(2)} \right]_{i_x,k_x=0}^{i_y,k_y=0\ldots Q-1} \qquad (13)$$

and can be used as $\mathbf{R}_p$ in (10).

## 3.2. The 2D Correlation Matrix Splitting Method

When texture is dynamic the correlation matrices (13) do not reflect dynamic nature of the image. And so the 2D correlation matrix (12) must be used for estimation of the texture resonance frequencies. A series of methods for the estimation of two-dimensional data resonance frequencies are known [24]. The following approaches to 2D EHD technique are developed this time: 2D-MUSIC (multiple signal classification) [30], 2D-ESPRIT (estimation of signal parameter via rotational invariance techniques) [31], 2D-matrices pencil method (MPM) [32] and its total least squares (TLS) version [33], 2D-linear prediction method (LPM) based on the 2D generalisation of the LSO (2) [34,35]. The MUSIC and ESPRIT use second data moments, the MPM uses first data moments, 2D LPM – first and second moments. The 2D EHD is based on the dynamic model of data first moments shift transforms along coordinates. We will consider a generalization of mentioned above approaches on the platform of SVD of correlation matrix like (12) and its splinter into pencils. The matrices pencils create eigenvalues problem that can be solved by TLS approach. The eigenvalues are equal to roots of polynomial (3). The eigenvalues problem may be written for 2D data $d_{m,n}$ of size $M \times N$ as matrices pencil [32,33]

$$\mathbf{D}_x^{(2)} - z_x \mathbf{D}^{(2)}; \quad \mathbf{D}_y^{(2)} - z_y \mathbf{D}^{(2)}, \quad (14)$$

where $\mathbf{D}_{x(y)}^{(2)} = \left[ \mathbf{D}_{i+k+x}^{(y)} \right]_{i=0...M-L-1}^{k=0...L-1}$,

$\mathbf{D}_{i+k+x}^{(y)} = \left[ d_{i+k+x, m+n+y} \right]_{n=0...N-L-1}^{m=0...L-1}$, $z$ – spectral parameter, $L$ – splitting parameter. If the 2D harmonics are not noised and its number $P \leq L$ then the rank of matrices pencil (14) is equal to $P$. If $z_x$ and $z_y$ are equal to matrices pencils (14) eigenvalues then expressions (14) are equal to zero. So, for the eigenvalues problem solution the following matrices can be used (for $x$ and $y$ coordinates).

$$\mathbf{Z}_{Ex(y)}^{(2)} = \left( \mathbf{D}^{(2)} \right)^{\#} \mathbf{D}_{x(y)}^{(2)}. \quad (15)$$

The SVD can be used for operation of pseudoinversion. The $P$ largest singular values and corresponded to them eigen vectors may be chosen. However in the case of image decomposition the SVD may be drastically time-consuming. Let's multiply matrices in (15) by matrix $\mathbf{D}^{(2)T}$ from the left. The products will be the correlation matrix $\mathbf{R}^{(2)} = \mathbf{D}^{(2)T} \mathbf{D}^{(2)}$ like (12) and its shifted parts. This operation can be presented using SVD of the matrix $\mathbf{R}^{(2)}$.

$$\mathbf{R}^{(2)} = \mathbf{U} \Xi \mathbf{V}^H, \quad (16)$$

where $\mathbf{U}$ and $\mathbf{V}$ – the matrices of the left and right orthogonal eigenvectors, $\Xi$ – the diagonal matrix of the singular values. Using expression (16) matrix (15) may be rewritten as

$$\mathbf{Z}_{Ex(y)}^{(2)} = \left( \mathbf{U}_0^H \mathbf{U}_0 \right)^{-1} \mathbf{U}_0^H \mathbf{U}_{x(y)}, \quad (17)$$

where matrices $\mathbf{U}_0$, $\mathbf{U}_{x(y)}$ – have $P$ vectors and that rows which correspond to reciprocally shifted matrices $\mathbf{D}^{(2)}$ and $\mathbf{D}_{x(y)}^{(2)}$. The matrix $\mathbf{U}_0$ is the basic in expression (17), the shifted matrix $\mathbf{U}_{x(y)}$ has structure that depends on direction of the shift: toward $x$ – $\mathbf{U}_x$; toward $y$ – $\mathbf{U}_y$. The basic matrix $\mathbf{U}_0$ is formed by extracting from matrix $\mathbf{U}$ last $M-L-1$ rows and each $N-L-1$ th row starting from the last row, also $P$ first vectors of the matrix $\mathbf{U}$ are using. The matrix $\mathbf{U}_x$ is forming by extracting from matrix $\mathbf{U}$ first $M-L-1$ rows and the matrix $\mathbf{U}_y$ by each $N-L-1$ th row starting from the first row. The inversion of the matrix $\mathbf{U}_0^H \mathbf{U}_0$ may be executed by iteration operation,

$$\mathbf{E}_k = \mathbf{E}_{k-1} + \frac{\left( \mathbf{E}_{k-1} \mathbf{u}_\alpha^H \right) \left( \mathbf{u}_\alpha \mathbf{E}_{k-1} \right)}{1 + \mathbf{u}_\alpha \mathbf{E}_{k-1} \mathbf{u}_\alpha^H}, \quad (18)$$

where $\mathbf{E}_0 = \mathbf{I}_P$ – the identity matrix of size $P$, $\mathbf{u}_\alpha$ – the $\alpha$ th row of matrix $\mathbf{U}$ first $P$ vectors, the indices $\alpha$ denote the set of rows that are extracting. If all extracting rows are used consequently in (18) then the last matrix $\mathbf{E}_k$ is equal to $(\mathbf{U}_0^H \mathbf{U}_0)^{-1}$. The eigenvalues $z_{xi}$, $z_{yi}$ ($i=1...P$) can be defined as eigenvalues of the matrices (17) modified depending on shift direction.

## 3.3. Texture Filtering

The textured image region of size $n_x \times n_y$ : $n_x > P$; $n_y > Q$, that covers texture kernel can be presented by the EHD like (8), its pixels values are the following.

$$d_{i,k} \approx \sum_{m=0}^{P-1} \sum_{n=0}^{Q-1} A_{m,n} z_{xm}^i z_{yn}^k, \quad (19)$$

where $i(k) = 0...n_{x(y)} - 1$, $A_{m,n}$ – spectral coefficients that are found by inverse to (19) transform like (6). The spectral coefficients of filtered image can be found as

$$E_{m,n} \approx \sum_{i=0}^{n_x-1} \sum_{k=0}^{n_y-1} E \, \bar{z}_{xm}^i \bar{z}_{yn}^k, \quad (20)$$

where $E$ – some constant value, $\tilde{z}_{xm}^i$ and $\tilde{z}_{yn}^k$ – elements of matrices $\mathbf{Z}^{\#}$ in (6). From (19) and (20) it follows that for obtaining of the flat signal the image spectrum must be transformed by the filter with spectrum

$$H_{m,n} = E_{m,n} A_{m,n}^{-1}. \quad (21)$$

The procedures of filtration in spectral domain by (6), (8) and (21) can be changed by filtration in spatial domain by the filter with the transient characteristic

$$h_{i,k} = \sum_{m=0}^{P} \sum_{n=0}^{Q} H_{m,n} \tilde{z}_{xm}^i \tilde{z}_{yn}^k. \quad (22)$$

This procedure seems as convolution operation

$$\sum_{m,n=0}^{P-1,Q-1} h_{m,n} d_{m+i,n+k} = E + \varsigma_{i,k}, \quad (23)$$

where $\varsigma_{i,j}$ – error noise.

The statistical analysis of the filtered signal was made using the dispersion of the noise of base region filtering. It was estimated as

$$\sigma_\varsigma^2 = \frac{1}{n_x n_y} \sum_{i,j=0}^{n_x-1, n_y-1} (\tilde{d}_{i,j} - E)^2, \quad (24)$$

where $\tilde{d}_{i,j}$ – filtered values of image signal. Texture pattern can be recognized using the condition

$$\mathbf{if} \bigcup_{j=1}^{3} \left( |\tilde{d}_{i,k}^{[t]} - E| > 3\sigma_\varsigma^{[t]} \right)$$

$$\mathbf{then}\ v_{i,k}^{[t]} = d_{i,k}^{[t]}\ \mathbf{else}\ v_{i,k}^{[t]} = 0;\ t=1,2,3\ , \quad (25)$$

where top indices in the brackets point on colors, the dispersion (24) was estimated for each color components, $d_{i,k}^{[t]}$ – initial image pixels values, $v_{i,k}^{[t]}$ – filtered image pixels values.

The filter (23) suppresses eigen or resonance harmonic modes of textured image signal. As it follows from expressions (21) and (22) the transient characteristic of the filter relates with inversion of texture spectrum and therefore the filter can be classified as inverse resonance filter.

## 4. Filtered Texture Abnormalities Analysis

The IRF, statistical (24) and logical analysis (25) recognize own texture, foreign objects and textures of another types. The filtering may gives errors because of texture heterogeneity. The errors removing can be executed by account of other image properties. We will consider two approaches that can be implemented for removing errors of static and dynamic textures filtering.

### 4.1. Histogram Difference Localization

Let's an foreign object was found and it was localized in square $S_o$ with coordinates $(x_0, y_0)$ and $(x_1, y_1)$. The extended square $S_e$ with coordinates $(x_0 - e, y_0 - e)$, $(x_1 + e, y_1 + e)$, where $e$ – some constant value $\sim 5...10$, can be defined. The gray scale difference histogram matrices can be defined as

$$\mathbf{G}_{row} = \left[\!\left[ g_{i,k} - g_{ek} \right]\!\right]_{i=x_0...x_1}^{k=1...K};$$

$$\mathbf{G}_{col} = \left[\!\left[ g_{k,j} - g_{ek} \right]\!\right]_{k=1...K}^{j=y_0...y_1}, \quad (26)$$

where $K$ – number of gray scales, $g_{i,\cdot}$, $g_{\cdot,j}$ – histogram functions along rows and columns of the $S_o$, $g_{e\cdot}$ – histogram of the $S_e - S_o$ region. The matrices (26) product gives the matrix

$$\mathbf{C} = \mathbf{G}_{row} \mathbf{G}_{col} \quad (27)$$

which shows space distribution of histogram functions difference. The binary matrix can be defined as

$$\mathbf{C}_b = \left[\mathbf{if}\ c_{i,k} > \varepsilon_c\ c_{b_{i,k}} = 1\ \mathbf{else}\ c_{b_{i,k}} = 0 \right]_{i=x_0...x_1}^{k=y_0...y_1}, (28)$$

where $\varepsilon_c$ – threshold level. If majority elements $c_{b_{i,k}} = 1$ of the matrix (27) are localized with some density or adjoining then the object is true else it is false. Also other criteria can be used for object state estimation, for example the value of weighed elements $c_{i,k} c_{b_{i,k}}$ sum. If image is color then three binary matrices (28) can be united by disjunction or conjunction operation.

### 4.2. Binary Correlation of Consecutive Image Frames

When texture is dynamic and characterized by nonstationary surges the application of filtration (23) and algorithm (25) does not provide high quality of the filtering and recognition. If texture tracking is continuous it is possible to remove false foreign objects by binary correlation of objects that are found in consequent frames.

The method of binary correlation filtration includes:
1) the objects size and geometrical centre estimation;

2) the correlation coefficient estimation using $L$ consecutive frames,

$$r = \frac{\sum_{t=0}^{L-1} \sum_{i=-Sx/2}^{Sx/2} \sum_{k=-Sy/2}^{Sy/2} \{v_{I+i,K+k}^{(0)} > 0\} \wedge \{v_{I+i,K+k}^{(t)} > 0\}}{\sum_{\tau=0}^{L-1} \sum_{i=-Sx/2}^{Sx/2} \sum_{k=-Sy/2}^{Sy/2} \{v_{I+i,K+k}^{(t)} > 0\}}, (29)$$

where $\{\cdot\}$ denotes the logical function, if condition in the brackets is true then it equal to one, else to zero, $I, K$ – the coordinates of the object centre of the frame number in top brackets, $Sx \times Sy$ – the object size.

3) if the correlation coefficient (29) exceeds threshold level then the object is true else false;
4) true object is redefined with surround background for further classification.

The $L$ frames running time must be much smaller in comparison with the time of target object moving outside of the rectangle of the size $Sx \times Sy$.

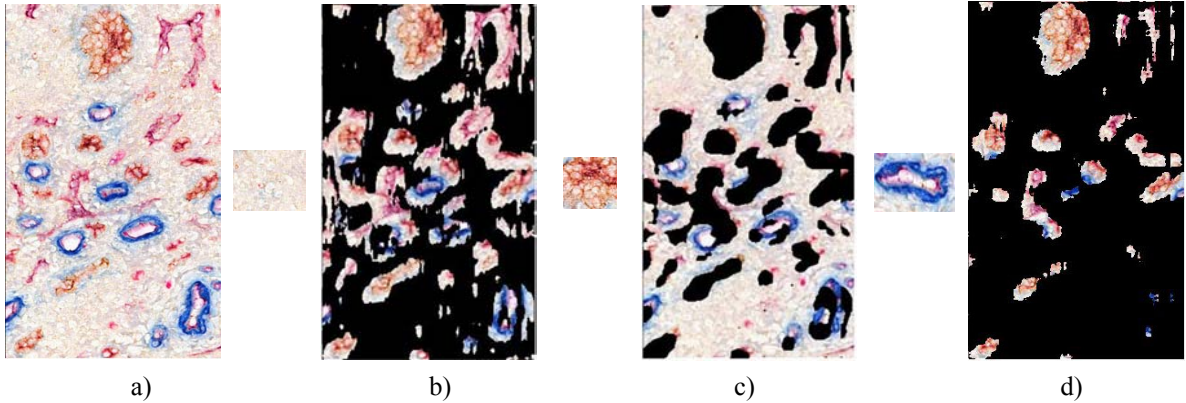

a) b) c) d)

Figure. 1. Initial image a) and three examples of pattern recognition b), c), d): filtering patterns and filtered images.

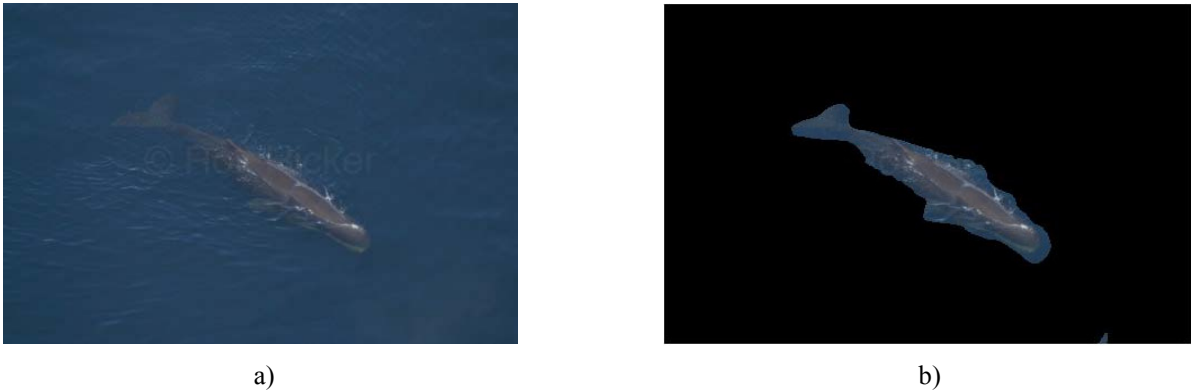

a) b)

Figure. 2. Initial image a) and filtered image b) with recognized object.

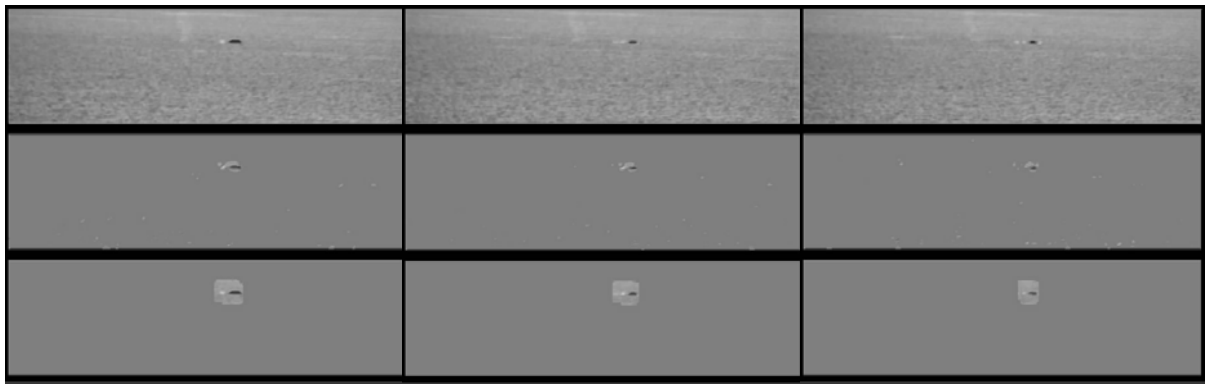

Figure 3. Three consecutive frames of marine surface filtering and mammal tracking: high positions – initial images; middle positions – images filtered by IRF; low positions – images filtered by correlative filter.

## 5. Filtering Experimental Tests

Figure 1 shows example of patterns recognition of biomedical nature image. Three types of patterns were used for inverse filters (23) of the order $16 \times 16$ design based on LS model (11) and correlation matrices (13). As it can be seen from figure the IRF traces heterogeneities boundaries, small and large foreign objects well. In the case of d) the combined pattern was filtered. Figure 2 shows example of object recognition on the dynamic background. The IRF order $16 \times 16$. The base region of size $64 \times 64$ in the left high image corner was used for the IRF design using the 2D correlation splintering method. The additional filtration was made by histogram difference analysis. The disjunction of three binary matrices (28) of three colour components was used for evaluation of histogram difference density and objects localisation. The difference density was evaluated by division of the object localization square on pixel cells of the size $5 \times 5$. The object is true if at least 0.75 of one cell is filled by elements equal to one. It is evident that the object visibility is not uniform but it was recognized fully. Three consecutive frames of marine surface filtering are shown in Figure 3 [36]. The base region of size $32 \times 32$ in left high image corner was used for design of the IRF of the order $8 \times 8$ using the 2D correlation splintering method (full frame size is $200 \times 600$). First frame of the series free from foreign objects was used for filter design. The binary correlative filter (29) with parameters $L=3$, $r=0.3$ was also used. As it follows from Figure 3 some mistakes of filtering by IRF were removed well by correlative filter.

## 6. Conclusions

Nowadays methods of texture analysis are found on the decompositions and integral transforms that have the same symmetry as texture in respect to shift and rotate transforms. The moving in some direction is natural for many types of static and dynamic textures. This type of transforms can be represented by linear shift operator that modeling step-type changes and simultaneously periodicity of the texture. The eigenvectors of LSO can serve as basis for EHD of the texture with the invariance to shift transforms.

The problem of pattern classification can be considered as inverse problem of recovering of known signal with flat surface. The dispersion value of the surface fluctuations points on recognized texture or foreign objects.

The IRF in conjunction with histogram difference in the case of static textures and correlative filtering in the case of dynamic textures series successfully removes textured signal and recognizes foreign objects boundaries.

Some peculiarities of textured image signals determine special claims to eigen harmonic decomposition method. Two approaches to this problem solution are offered. Comparative study of the methods shows that they give similar results when image is regular or quasi-regular. In the case of dynamic texture the 2D splitting method gives better results and this is natural because it uses full 2D correlation matrix and needs in much more operation. But if it gives smaller number of filtering errors then its implementation is expedient because the procedure of errors removing is not so simple as convolution operation (23). The real time tracing of dynamic texture series usually needs in the filter redefinition ones per several minutes and so it is enough time for filter parameters evaluation by the 2D method. The dumping factors of matrices (17) eigenvalues usually reflect well the dynamic properties of signal and in the case of textured image they are in neighbourhood of zero and so the filter suppresses texture periodicity well.

The considered above method of textured image filtering for foreign objects recognizing or textures classification differs from other known methods by the following features:
- simplicity – each texture is characterized by its own filter and dispersion of the error;
- ability of implementation in real time using arithmetic units array;
- regulated resolution and quality;
- invariance to image moving changes;
- it is based on principal harmonic decomposition that is simpler and more convenient than ICA, EMD or EVD because has simple analytical form and may be simply adapted to image and its fragments size.